\documentclass{article}

\usepackage[final]{corl_2024} 
\usepackage{graphicx}
\usepackage{amsmath}
\usepackage{subcaption}   
\usepackage{multirow}
\usepackage{enumitem}
\usepackage{booktabs}

\title{Traffic and Safety Rule Compliance of Humans in Diverse Driving Situations}

\author{
  Michael Kurenkov$^{1,2,}$\thanks{These authors contributed equally to this work.} , Sajad Marvi$^{1,3,*}$, Julian Schmidt$^1$, Christoph B. Rist$^1$,\\ \textbf{Alessandro Canevaro$^1$, Hang Yu$^1$, Julian Jordan$^1$, Georg Schildbach$^2$, Abhinav Valada$^3$}\\   
$^1$ Department of Automated Driving Software Functions, Mercedes Benz AG, Germany\\
$^2$ Department of Electrical Engineering in Medicine, University of Lübeck, Germany\\
$^3$ Department of Computer Science, University of Freiburg, Germany
}

\begin{document}
\maketitle


\begin{abstract}
    The increasing interest in autonomous driving systems has highlighted the need for an in-depth analysis of human driving behavior in diverse scenarios. Analyzing human data is crucial for developing autonomous systems that can replicate safe driving practices and ensure seamless integration into human-dominated environments. This paper presents a comparative evaluation of human compliance with traffic and safety rules across multiple trajectory prediction datasets, including Argoverse 2, nuPlan, Lyft, and DeepUrban. By defining and leveraging existing safety and behavior-related metrics such as time to collision, adherence to speed limits, and interactions with other traffic participants, we aim to provide a comprehensive understanding of each dataset's strengths and limitations. Our analysis focuses on the distribution of data samples, identifying noise, outliers, and undesirable behaviors exhibited by human drivers in both the training and validation sets. The results underscore the need for applying robust filtering techniques to certain datasets due to high levels of noise and the presence of such undesirable behaviors.
\end{abstract}

\keywords{Autonomous Driving, Dataset Analysis} 


\section{Introduction}
	
    In recent years, autonomous vehicles (AVs) have gained significant attention due to their potential to reduce traffic fatalities. The widespread adoption of AV technology is contingent not only on technical performance but also on public trust, with concerns centering on safety and potential technological malfunctions~\cite{forbes_survey, londono2024fairness}. A key factor in improving trust in autonomous systems is the ability to understand and replicate human driving behavior. However, worldwide, road accidents cause over 1.19 million deaths annually, with a majority resulting from human error~\cite{WHO_Global_Road_Safety_2023}, hence following human driving pattern is not always desired. Since the majority of accidents are caused by human error, analyzing human driving data allows us to identify common mistakes and undesirable driving patterns. This understanding is crucial for training machine learning models, such as those used in behavior cloning, where the goal is to mimic human driving behavior.
    Identifying undesirable driving patterns is especially useful for achieving a defensive driving behavior, which is proven to play a significant role in increasing passenger comfort and trust in AVs~\cite{Hartwich_Hollander_Johannmeyer_Krems_2021}. 
    

In summary, our main contributions are:
\begin{itemize}[topsep=0em,noitemsep]
\item Review existing criticality measures and safety violations and define new relevant metrics
\item Provide comprehensive overview of dataset characteristics
\item Examine critical behavior across datasets and compare findings
\end{itemize}


\section{Related Work}
\label{sec:related_work}
The following sections will provide a summary of the current state of the art in metrics, datasets, and trajectory prediction methodologies.

\begin{figure}[t]
    \centering
    \begin{subfigure}[b]{0.60\textwidth}
        \centering
        \includegraphics[width=\textwidth]{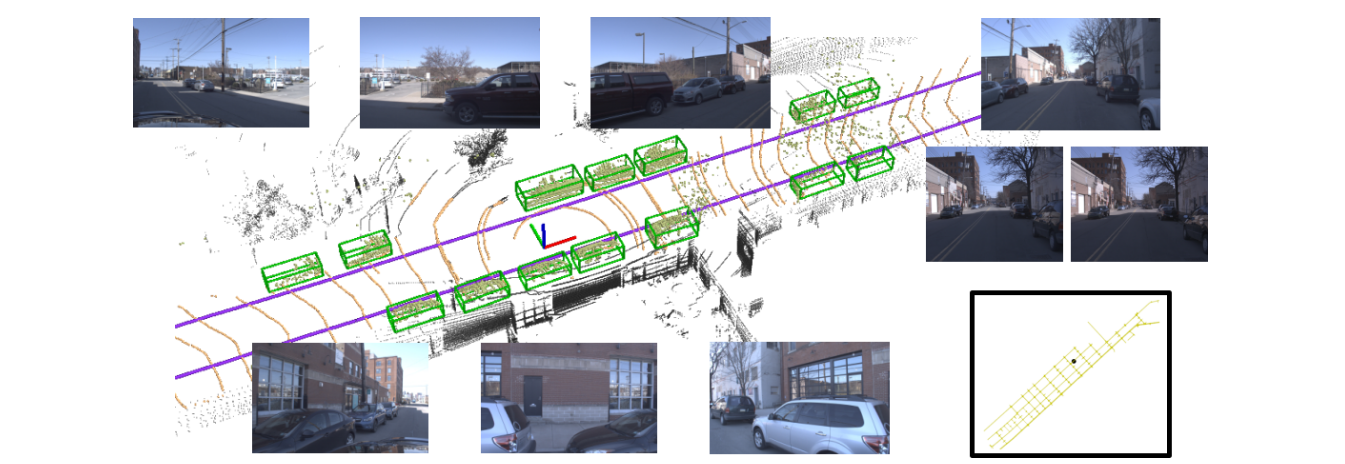}
        \caption*{(a)}  
    \end{subfigure}
    \hfill
    \begin{subfigure}[b]{0.39\textwidth}
        \centering
        \includegraphics[width=\textwidth]{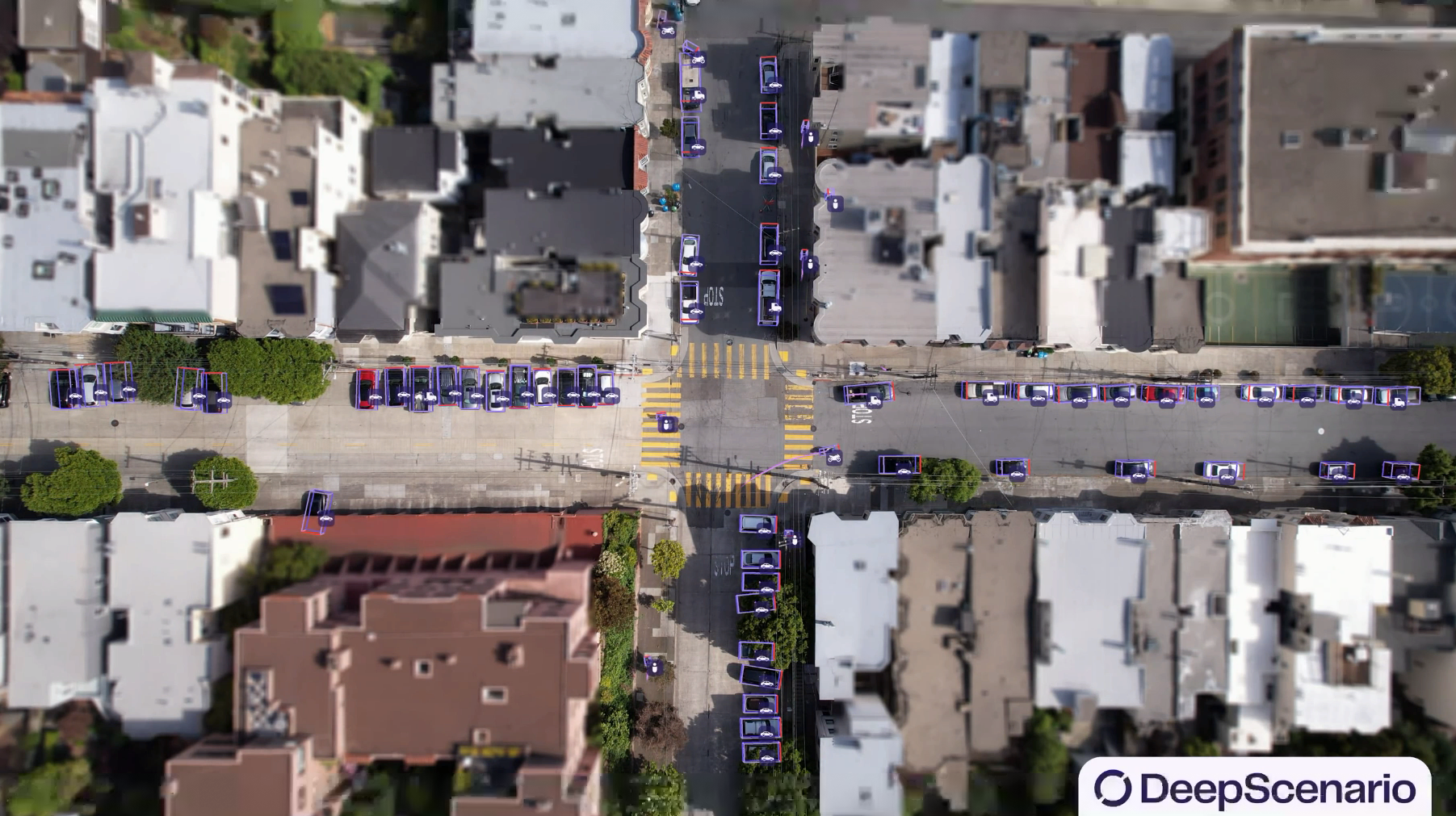}
        \caption*{(b)}  
    \end{subfigure}
    \caption{Example of an onboard sensor dataset sample from (a)~Argoverse2~\cite{wilsonArgoverseNextGeneration2023a} and (b)~drone data DeepUrban~\cite{deepurban2024}.}
    \label{av2}
    \label{deepscenario}
    \vspace{-0.25cm}
\end{figure}

\subsection{Datasets for Trajectory Prediction and Planning}\label{AA}
This section explores key datasets containing human-driving data.
The quality of machine learning models trained using these datasets (e.g., trajectory prediction models or planner models) relies heavily on the driving behavior and patterns demonstrated in these datasets. 
Early projects such as the 100-Car Naturalistic Study~\cite{nealeOverview100carNaturalistic2005} collected large amounts of data but were primarily suited for perception tasks. More recent datasets are tailored for trajectory prediction, directly providing data on agent and lane level instead of raw sensor data only (Table~\ref{tab:datasets}).

\textbf{Onboard Sensor Datasets}: 
These datasets use multiple sensors (e.g., cameras, LiDAR, Radar) from the ego vehicle to detect its surroundings, capturing real-time measurements like speed and acceleration (Fig.~\ref{av2}(a)). However, onboard sensors are prone to noise and limited in their ability to capture distant traffic conditions. Despite these limitations, onboard sensor data is essential for the full AV stack, including trajectory prediction, planning, perception, and control, due to its ability to provide highly accurate internal vehicle signals.

\textbf{External Sensor Datasets}: 
External datasets, recorded via drones or elevated cameras, offer a top-down view of traffic as shown in Fig.~\ref{deepscenario}(b). These methods offer an alternative perspective by precisely tracking all traffic participants, including those not visible from the ego vehicle’s viewpoint. However, external sensor datasets face limitations such as restricted availability (strict flight regulations) and environmental dependencies (weather and lighting conditions).

\begin{table}[t]
\centering
\footnotesize
\begin{tabular}{l|c c c c c}
\toprule
\textbf{Dataset} & \textbf{Cities} & \textbf{Recording Type} & \textbf{Total Volume (h)} & \textbf{Year} & \textbf{Access} \\
\midrule
Argoverse 1~\cite{changArgoverse3DTracking2019} & 2 & Onboard & 320 &  2019 & Open \\ 
Lyft L5~\cite{DBLP:journals/corr/abs-2006-14480} & 1 & Onboard & 1118 &  2020 & Open\\ 
Waymo~\cite{ettingerLargeScaleInteractive2021} & 6 & Onboard & 574 &  2021 & Open\\
Shifts~\cite{malininShiftsDatasetReal2022} & - & Onboard & 1667 &  2021 & Open\\ 
nuPlan~\cite{caesarNuPlanClosedloopMLbased2022} & 4 & Onboard & 1500 &  2021 & Open\\ 
Argoverse 2~\cite{wilsonArgoverseNextGeneration2023a} & 6 & Onboard & 763 &  2023 & Open\\
\midrule
highD~\cite{highDdataset} & - & Drone & 16 &  2018 & Open\\
INTERACTION~\cite{zhanINTERACTIONDatasetINTERnational2019} & - & Drone & 16 & 2019 & Open\\
inD~\cite{inDdataset} & - & Drone & 10 &  2020 & Open\\
MONA~\cite{gressenbuchMONAMunichMotion2022} & - & Elevated Camera & 130 &  2022 & Open\\ 
DeepUrban~\cite{deepurban2024} & - & Drone & 66 &  2024 & Restricted\\
\bottomrule
\end{tabular}
\vspace{5pt}
\caption{Naturalistic Driving Studies collection over the years.}
\label{tab:datasets}
\vspace{-0.5cm}
\end{table}

\subsection{Criticality Measures and Traffic Rule / Safety Violations}

Ensuring and demonstrating safe driving behavior is a crucial step for the widespread adoption of AVs, particularly for the certification of SAE Level 5 systems~\cite{J3016_202104TaxonomyDefinitions}, where human intervention is not possible. Both the research and industrial sectors are keen on identifying critical driving situations and maneuvers~\cite{schmidt2022meat} and ensuring appropriate responses.

\textbf{Criticality Measures}: 
Criticality measures are essential for analyzing human driving behavior, helping to quantify the risk involved in specific traffic scenarios. Several key metrics are commonly used for this purpose, including Gap, Time to Collision (TTC), and Post Encroachment Time (PET)~\cite{westhofenCriticalityMetricsAutomated2023}.

\textit{Gap}: The Gap metric measures the distance  between two agents \( A_i \) and \( A_j \) at a given time \( t \). In a car following scenario, this is described as:
\begin{equation}
\label{GAP_eq}
\text{GAP}(A_i, A_j, t) = \text{distance}(p_{A_i,\text{front}}(t), p_{A_j,\text{rear}}(t)),
\end{equation}
where $p_{A_1,\text{front}}(t)$ is the 2D position of the front of agent $i$ at time step $t$.
The Headway (HW) is a related measure that calculates the distance between the fronts of two agents~\cite{haywardNEARMISSDETERMINATIONUSE1972}.

\textit{Time to Collision (TTC)}: TTC estimates the time until two agents would collide given a dynamic motion model at the current time step~\cite{haywardNEARMISSDETERMINATIONUSE1972, westhofenCriticalityMetricsAutomated2023}. 
It is widely used as a surrogate safety indicator. The critical threshold for TTC, \( \text{TTC}^* \), varies across the literature, with suggested values ranging from 1.5 to 4 seconds~\cite{vanderhorstTimetocollisionCueDecisionmaking1991, naseralaviGeneralFormulationTimetocollision2013, deveauxExtractionRiskKnowledge2021}.
It is calculated as
\begin{equation}
\label{ttc_eq}
\text{TTC}(A_i, A_j, t) = \min \left( \{ \tilde{t} \geq 0 \mid d(p_{A_i}(t + \tilde{t}), p_{A_j}(t + \tilde{t})) = 0 \} \cup \{\infty\} \right).
\end{equation}

There are also different variations of TTC, which have been used for evaluating the quality of planned trajectory. One variation used in CommonRoad~\cite{linCommonRoadCriMeToolboxCriticality2023} takes into consideration the relative velocity, acceleration, and Gap in the car following the scenario.

\textit{Post Encroachment Time (PET)}: PET calculates the time interval between one agent exiting and another agent entering a designated conflict area (CA):
\begin{equation}
\text{PET}(A_i, A_j, \text{CA}) = t_{\text{entry}}(A_j, \text{CA}) - t_{\text{exit}}(A_i, \text{CA}).
\end{equation}
PET can also be semi-predictive, with \( t_{\text{entry}}(A_j, \text{CA}) \) being predicted using motion models such as constant velocity.

\textbf{Traffic Rule Violations}: 
Legal compliance is a critical aspect of analyzing Naturalistic Driving Studies, as most accidents are caused by human error~\cite{Singh_2018}. While criticality measures may detect potential risks, they do not always capture traffic rule violations, which can also increase the risk of accidents. AVs must adhere strictly to traffic regulations, and defining such rules precisely for autonomous systems is essential. Driver errors and rule violations can be classified into three categories: recognition, decision, and reaction errors, with most accidents caused by the first two~\cite{khattakTaxonomyDrivingErrors2021}. In this study, the focus is on analyzing rule violations, particularly according to German law (StVO)~\cite{StVO2013Nichtamtliches}. Key traffic rules examined include:
\begin{itemize}[topsep=0pt,itemsep=0pt]
    \item Boundaries: Prohibition of crossing solid lines (StVO Section 2).
    \item Overtaking: Maintaining a minimum distance of 1.5~meters when overtaking pedestrians or cyclists in urban areas and 2~meters outside urban areas (StVO Section 5).
    \item Pedestrian Crossings: Vehicles must stop for pedestrians at crossings and must not block the crossing during congestion (StVO Section 26).
\end{itemize}
By examining such violations in motion prediction datasets, this analysis helps ensure that AVs comply with traffic regulations to reduce the risk of accidents.


\section{Critical Behavior Extraction Method}
\label{sec:critical_behavior_extraction_method}
In this section, first the various metrics used for dataset analysis are discussed. The second part focuses on the implementation of the metrics and details of the dataset analysis.

\subsection{Definition of Criticality Measures and Safety Rule Violations}

The analysis of human driving behavior, particularly in the context of autonomous driving, requires specific criteria that are generally applicable to all datasets.
The following criteria were used in this work:

\textit{Speed (VEL)}: 
One of the fundamental criteria is the speed of vehicles, which can be calculated from the agents' positions:
\begin{equation}
    \text{VEL} = \frac{p_{A_i}(t+1) - p_{A_i}(t)}{\Delta t},
\end{equation}
where \( p_{A_i}(t) \) is the position of agent $i$ at time \( t \), and \( \Delta t \) is the time step. For this analysis, speeds above 14 m/s (around 50 km/h) are considered critical, based on urban area speed limit
regulations of the US (30mph $\sim$ 48kph) and Germany (50kph).

\textit{Acceleration (ACC)}: 
Acceleration is another key criterion computed from the vehicle's speed:
\begin{equation}
\text{ACC} = \frac{v_{A_i}(t+1) - v_{A_i}(t)}{\Delta t},
\end{equation}
where $v_{A_i}(t)$ is velocity of the agent $i$ at time $t$. Acceleration values below \(-6 m/s^2\) and above \(6 m/s^2\) are considered critical, as research has shown that such values can cause discomfort or even whiplash~\cite{maselloUsingContextualData2023}.

\begin{figure}[t]
\centering
    \begin{subfigure}[b]{0.48\linewidth}
        \centering
        \includegraphics[width=\linewidth]{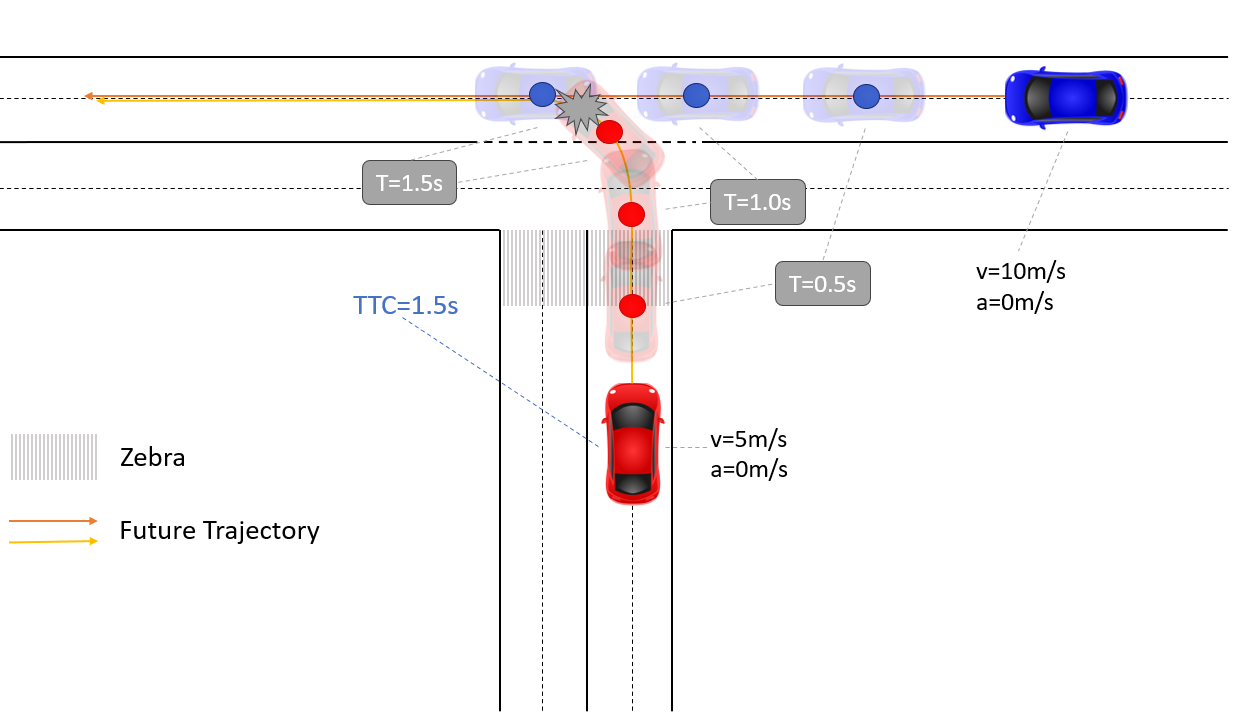}
        \caption*{(a)}  
        \label{subfig:ttc}
    \end{subfigure}
    \hspace{0.01\linewidth}
    \begin{subfigure}[b]{0.48\linewidth}
        \centering
        \includegraphics[width=\linewidth]{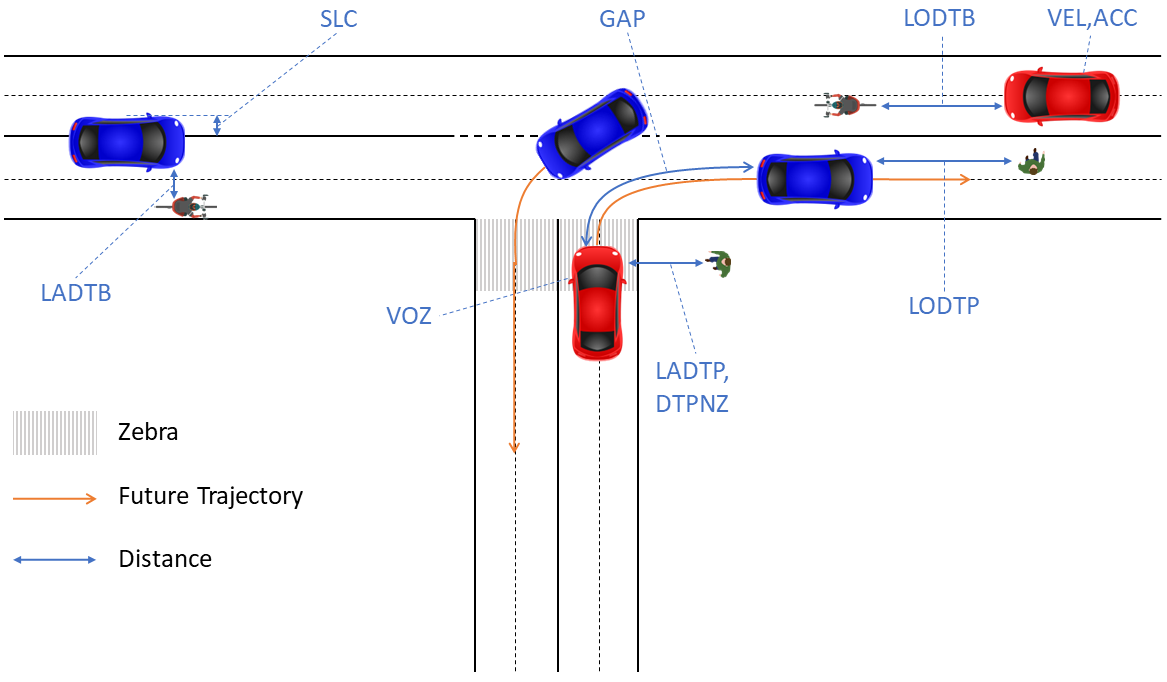}
        \caption*{(b)}  
        \label{subfig:metric_vis}
    \end{subfigure}

\caption{(a) TTC calculation where the red vehicle turns left and the blue vehicle follows the lane, using constant acceleration for future trajectory. (b) Visualization of all metrics except TTC in our framework.}
\label{fig:ttc_vis}
    \vspace{-0.2cm}
\end{figure}

\textit{Gap}: 
The calculation of the Gap is based on Eq.\ref{GAP_eq}. We consider the trajectory of the ego as a reference, and if other agents' polygon at each time step are positioned in the future trajectory of the ego, the minimal distance between two agents along the trajectory is considered as GAP value (Fig.~\ref{fig:ttc_vis}~(b)). 

\textit{Time to Collision (TTC)}: 
TTC is a key criterion used to assess how long it will take for two vehicles to collide if their current trajectories are maintained. The general formula for TTC is shown in Equation~\ref{ttc_eq}.
It calculates the minimum time \( t_\text{ttc} \) until a potential collision between two vehicles, \( A_i \) and \( A_j \), based on their current trajectories.
To make this calculation more general and not just limited to car-following scenarios, the vehicles' future trajectories are used as the reference paths. At each discrete time step \( t_\text{ttc} \in \{0.5, 1.0, 1.5, \dots, 39.5, 40\} \), the predicted positions \( \hat{p}_{A_{i}}(t + t_\text{ttc}), \hat{p}_{A_{j}}(t + t_\text{ttc}) \) for vehicles \( A_{i}, A_{j} \) are computed based on their current speed \( v_t \) and acceleration \( a_t \) by interpolating the traveled distance:
\[
\Delta X = v_t \cdot t_\text{ttc} + \frac{1}{2} \cdot a_t \cdot t_\text{ttc}^2
\]
over the reference path. The TTC value is determined as the first time step when the vehicles occupy the same position, indicating a potential collision.
(Fig.~\ref{fig:ttc_vis}(a)).
\begin{equation}
\text{TTC} = min(\{t_\text{ttc}| \hat{p}_{A_{i}}(t + t_\text{ttc})\cap \hat{p}_{A_{j}}(t +t_\text{ttc})\}), \text{ for } t_\text{ttc} \in \{0.5, 1.0, 1.5, \dots, 39.5, 40\}\text{s},
\end{equation}
where $\hat{p}$ is the function that outputs the interpolated future position of the agent.
Critical TTC thresholds typically vary depending on environmental factors and road conditions. Values in the range of 1.5 to 4~seconds have been suggested in the literature~\cite{benmimoun2014detection,iso2013intelligent}. For this study, a threshold of 2 seconds is chosen to ensure that only the most critical situations are flagged. 

\textit{Distance to Bicycles (DTB)}: 
The Distance to Bicycles (DTB) criterion is used to ensure safe interactions between vehicles and cyclists. The distance from a vehicle \( A \) to a cyclist \( C \) is calculated at each time step as
\begin{equation}
\text{DTB} = \text{distance}(p_{A}(t), p_{C}(t)).
\end{equation}
This analysis focuses on values within a 5 meter radius. To address specific legal requirements, such as the StVO in Germany, DTB is further divided into longitudinal and lateral components. The longitudinal DTB (LODTB) measures the forward or backward distance along the vehicle’s direction of travel relative to the cyclist, while the lateral DTB (LADTB) assesses the lateral distance between the vehicle and cyclist.

A minimum lateral distance of 1.5~meters must be maintained during overtaking in urban areas.

\textit{Distance to Pedestrians (DTP)}: 
Similar to the DTB, the Distance to Pedestrians (DTP) criterion measures the distance between a vehicle \( A \) and pedestrians \( P \) as
\begin{equation}
    \text{DTP} = \text{distance}(p_{A}(t), p_{P}(t)).
\end{equation}
As with the DTB, this distance is evaluated within a 5~meter range and divided into longitudinal (LODTP) and lateral (LADTP) components to ensure compliance with pedestrian safety regulations. The critical threshold of 1.5 meters is again used as the minimum distance vehicles should maintain from pedestrians in urban areas.




\textit{Pedestrian Crossing (DTPNZ)}: 
The DTPNZ criterion, a subset of the DTP, measures the proximity of pedestrians to the vehicle when it crosses a pedestrian zone. The criterion, therefore, gives special attention to whether the vehicle complies with traffic rules when approaching crosswalks. The critical threshold of 1.5~meters applies here as well, ensuring pedestrian safety.

\textit{Speed at Pedestrian Crossings (VOZ)}: 
This criterion measures vehicle speed when crossing pedestrian zones, derived from the general speed criterion (VEL)~\cite{varhelyiDriversSpeedBehaviour1998, budzynskiAssessmentInfluenceRoad2021}. Speeds are analyzed to ensure they comply with legal speed limits and safety regulations near crosswalks.

\textit{Solid Line Crossing (SLC)}: 
Crossing solid lines is a common traffic violation. The distance between the vehicle's outer tire and the solid line is measured as
\begin{equation}
\text{SLC} = \text{distance}(\text{outer\_tire}(A_t), \text{solid\_line}).
\end{equation}
This criterion helps assess whether the vehicle complies with lane discipline rules, as crossing a solid line can increase the risk of collisions. All the investigated criteria and their corresponding critical thresholds are aggregated in Table~\ref{tab:metrics}.

\begin{table}[t]
\centering
\footnotesize
\begin{tabular}{l|cccc}
\toprule
Name & Category & Unit & Requires Map & Criticality Threshold \\
\midrule
VEL & Traffic Violation & $m/s$ & No & $ \text{VEL}>14~m/s $  \\ 
ACC & Criticality Measure & $m/s^2$ & No   & $ \lvert\text{ACC}\rvert>6~m/s^2 $   \\ 
GAP & Criticality Measure & $m$ & No & -  \\ 
TTC & Criticality Measure & $s$ & No  & $\text{TTC}<2~s$  \\ 
DTB & Traffic Violation & $m$ & No  & $\text{DTB}<1.5~m$  \\ 
DTP & Traffic Violation & $m$ & No  & $\text{DTP}<1.5~m$  \\ 
DTPNZ & Traffic Violation & $m$ & Yes  & $\text{DTPNZ}<1.5m~$  \\ 
VOZ & Criticality Measure & $m/s$ & Yes  & -  \\ 
SLC & Traffic Violation & $m$ & Yes  & $\text{SLC}>0~m$  \\ 
\bottomrule
\end{tabular}
\vspace{5pt}
\caption{Metrics that are analyzed in the work with criticality threshold.}
\label{tab:metrics}
\vspace{-0.5cm}

\end{table}

\subsection{Analysis of Datasets}

The analysis of various datasets is a central component of this work. To make the evaluation process more efficient, we used a common internal representation, similar to TrajData~\cite{fengUniTrajUnifiedFramework2024} of the traffic scene in our problem formulation.
The representation contains key data required for safe autonomous driving, such as:
\begin{itemize}[topsep=0pt,itemsep=0pt]
\item \textit{Map data:} Contains information about lane borders, lane types (e.g., bus, bicycle lanes), boundaries (e.g., dashed, solid lines), crosswalks, and restricted areas. A graph-like structure links lanes via predecessors, successors, and neighbors.
\item \textit{Agent data:} Tracks the ego vehicle and other agents (e.g., buses, bicycles, pedestrians), including their position, speed, dimensions, and categories.
\item \textit{General data:} Stores scenario metadata such as the recording city, dataset name, frame rate, and time of day.
\end{itemize}

We standardized scenarios to 11 seconds at 10 frames per second, including only relevant map regions within 150 meters of the agents. The framework classifies agents and scenarios as critical or non-critical based on metrics and thresholds, filtering out undesirable behaviors for downstream tasks. It also generates metadata on critical agents or scenarios, as shown in Table~\ref{tab:criticalagent}.


 \begin{table}[t]
    \centering
    \footnotesize
    \begin{tabular}{l|cc|cc|cc|c}
    \toprule
    \multirow{2}{*}{\textbf{Metric}} & \multicolumn{2}{c|}{\textbf{Lyft}} & \multicolumn{2}{c|}{\textbf{nuPlan}} & \multicolumn{2}{c|}{\textbf{Argoverse 2}} & \multirow{2}{*}{\textbf{DeepUrban}} \\ 
     & Train & Val & Train & Val & Train & Val & \\
     \midrule
    VEL (${m}{s^{-1}}$)& 5.78 & 5.8 & 7.38 & 6.51 & 7.11 & 7.13 & 6.39 \\ 
    ACC (${m}{s^{-2}}$)& 0.11 & 0.11& 0.14 & 0.14 & 0.14 & 0.14 & 0.2 \\ 
    GAP ($m$)&  25.84 & 25.74 & 20.53 & 19.45 & 30.10 & 20.60 & 18.40 \\ 
    TTC  ($s$)& 6.0 & 6.0 & 7.0 & 7.0 & 6.0 & 6.0 & 7.0  \\ 
    LADTB  ($m$)& 2.13 & 2.04 & 1.75 & 1.86 & 1.91 & 1.69 & 2.03 \\ 
    LODTB  ($m$)& 7.17 & 7.81 & 7.83 & 7.42 & 7.17 & 6.69 & 7.46  \\ 
    LADTP  ($m$)&  2.4 & 2.39 & 2.35 & 2.32 & 2.16 & 2.15 & 2.18 \\ 
    LODTP  ($m$)& 0.98 & 1.81 & 3.68 & 3.64 & 3.75 & 2.54 & 3.68\\ 
    DTPNZ  ($m$)& 3.59 & 3.52 & 2.61 & 3.11 & 3.07 & 3.0 & 3.36  \\ 
    VOZ (${m}{s^{-1}}$)& 3.86 & 3.82 & 4.45 & 4.5 & 4.96 & 5.0 & 2.88 \\ 
    SLC  ($m$)& 0.23 & 0.22 & 0.20 & 0.24 & 0.22 & 0.21 & 0.11 \\
    \bottomrule
    \end{tabular}
    \vspace{5pt}
    \caption{Comparison of median values for every metric and dataset.}
        \vspace{-0.5cm}
    \label{tab:medians}
    
\end{table}

\begin{figure}[htpb]
    \centering
    \begin{subfigure}[b]{0.45\textwidth}
        \centering
        \includegraphics[width=\textwidth]{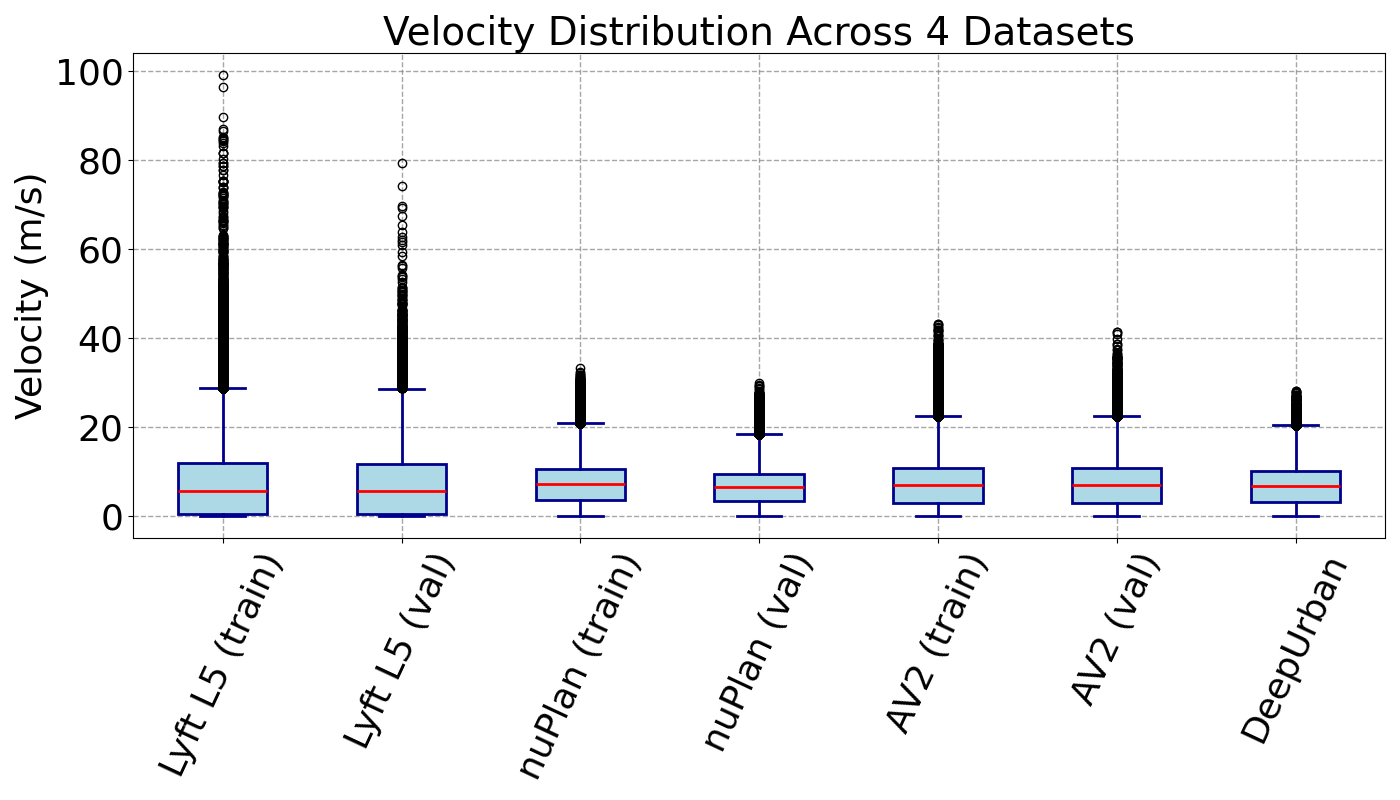}
    \end{subfigure}
    \hfill
    \begin{subfigure}[b]{0.45\textwidth}
        \centering
        \includegraphics[width=\textwidth]{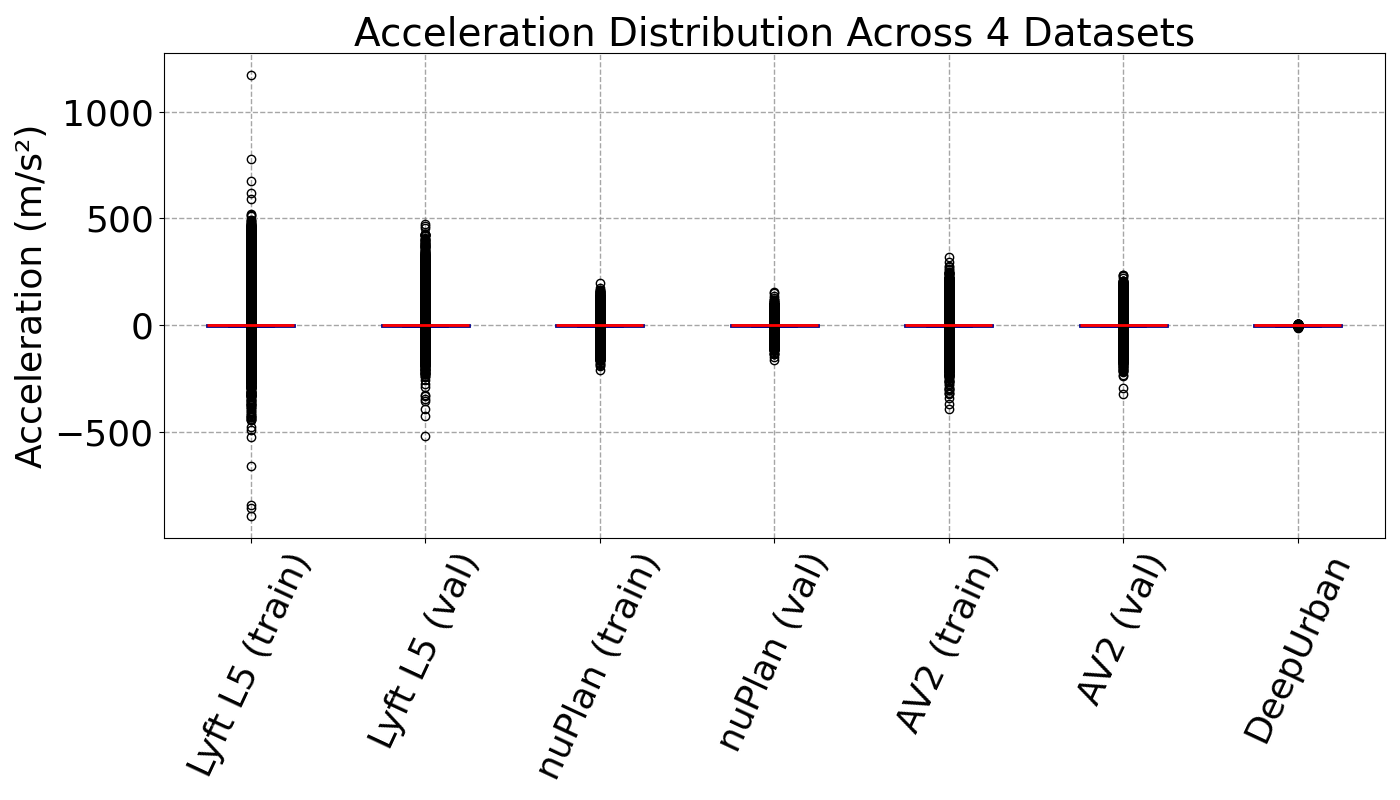}
    \end{subfigure}

    \begin{subfigure}[b]{0.45\textwidth}
        \centering
        \includegraphics[width=\textwidth]{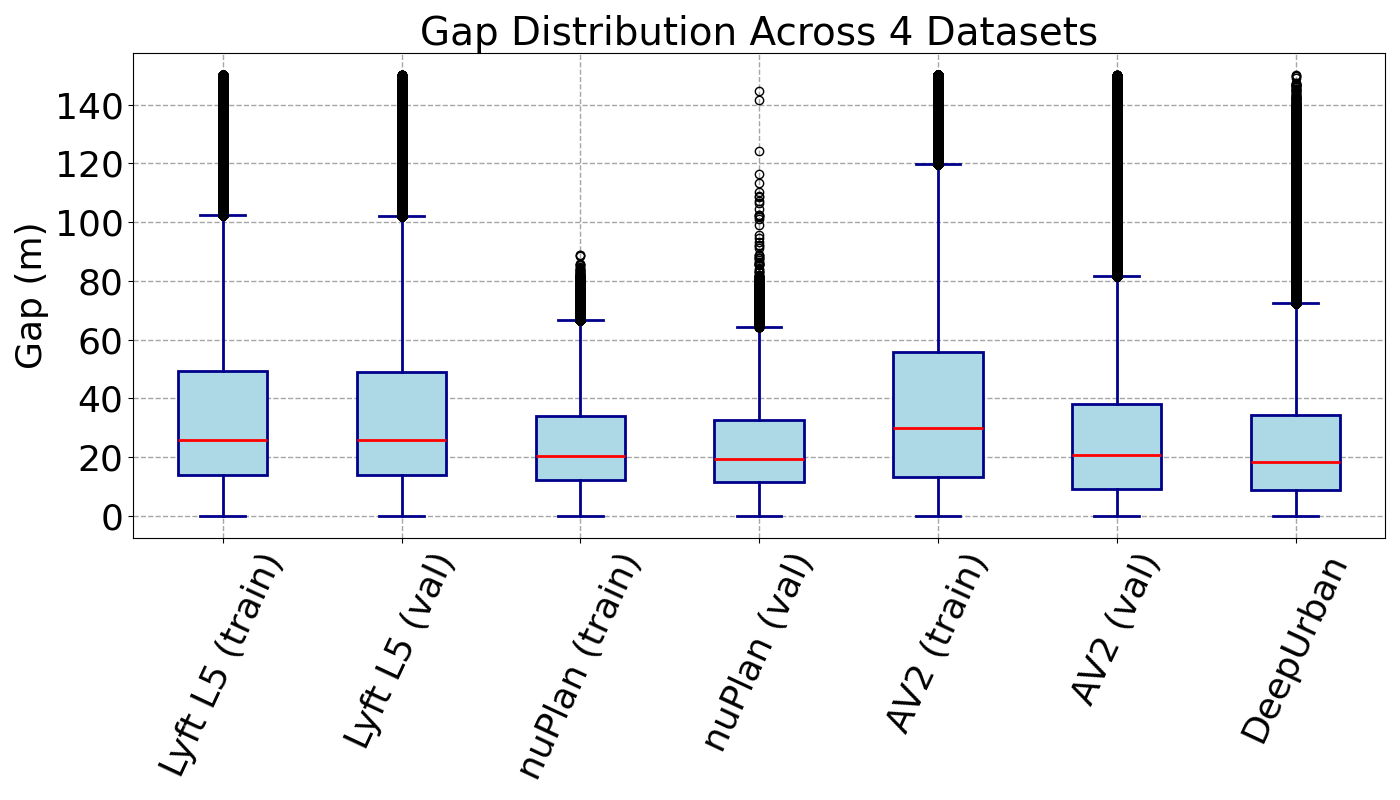}
    \end{subfigure}
    \hfill
    \begin{subfigure}[b]{0.45\textwidth}
        \centering
        \includegraphics[width=\textwidth]{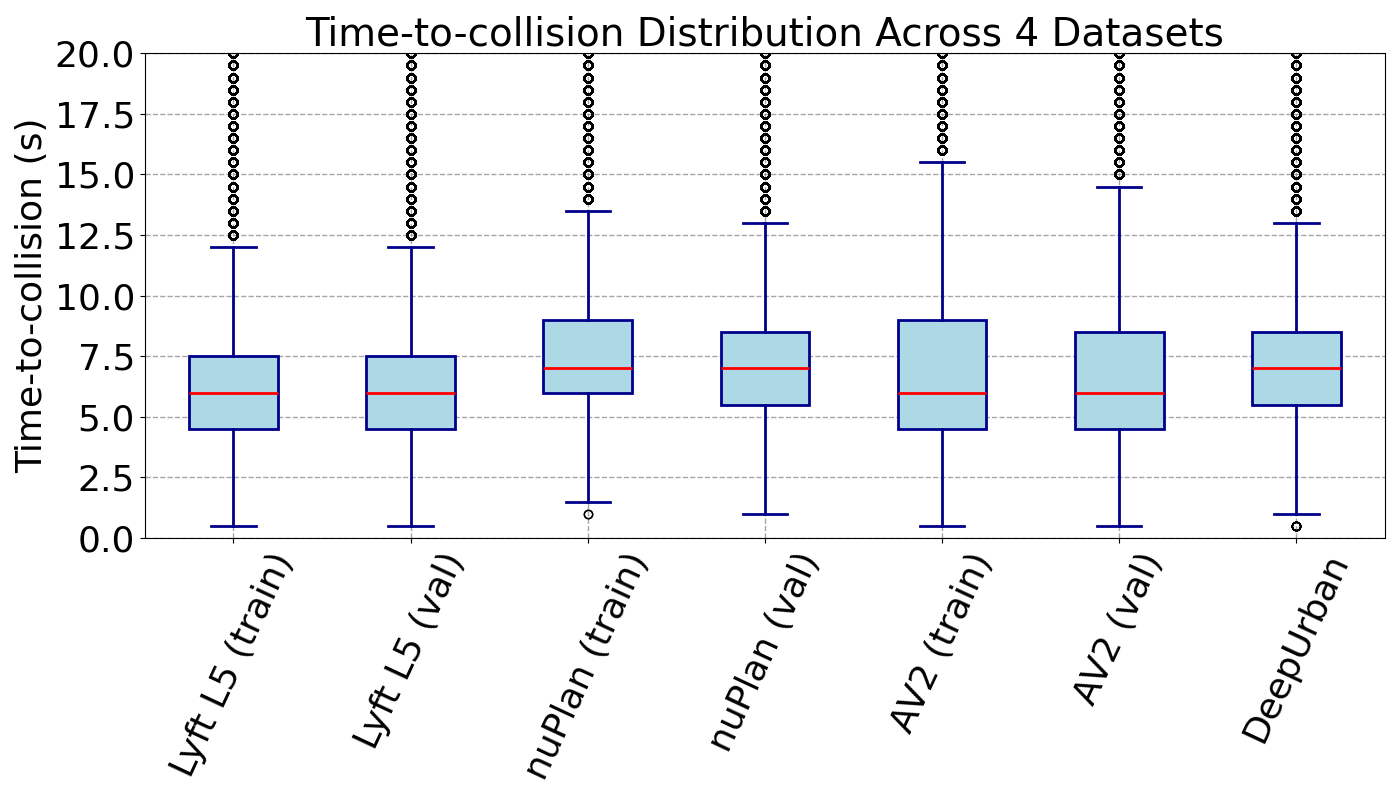}
    \end{subfigure}

    \begin{subfigure}[b]{0.45\textwidth}
        \centering
        \includegraphics[width=\textwidth]{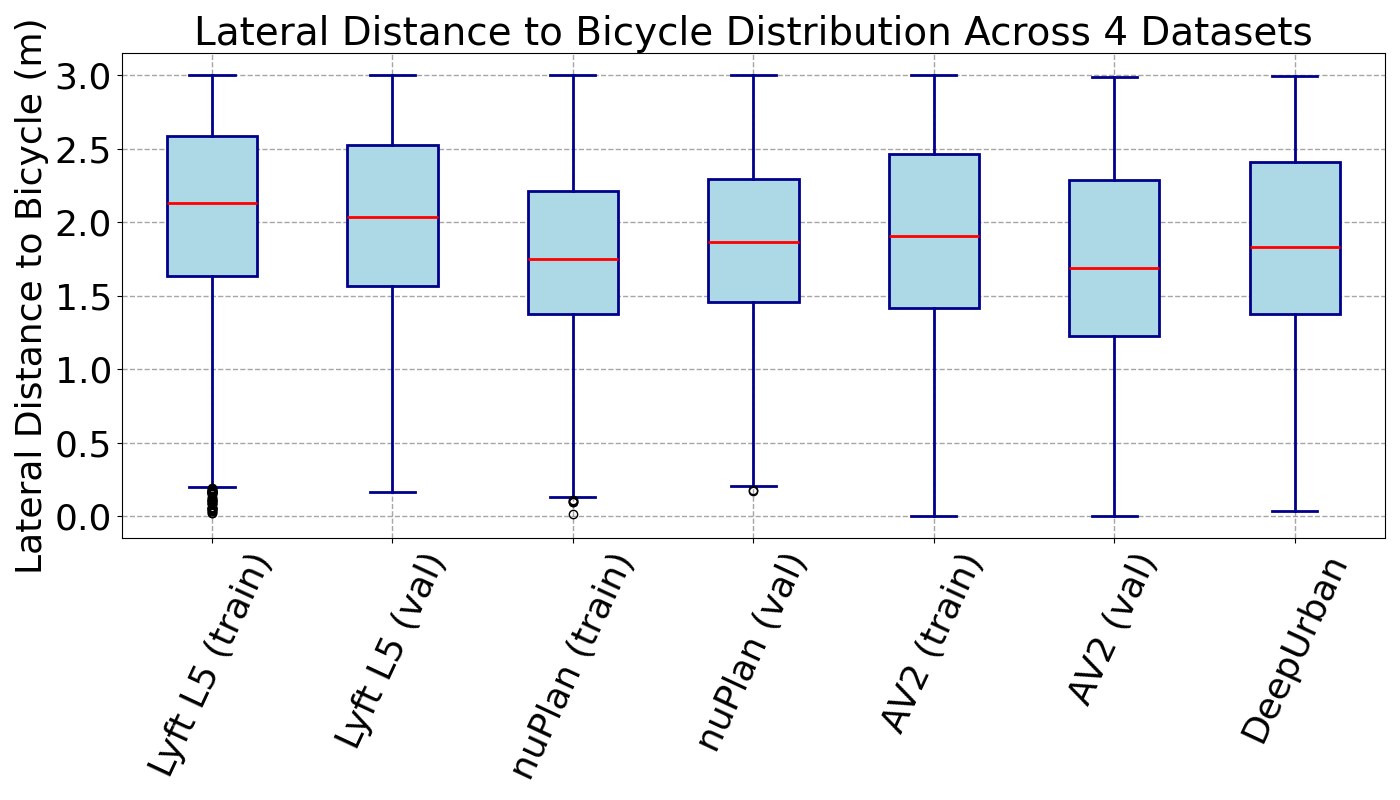}
    \end{subfigure}
    \hfill
    \begin{subfigure}[b]{0.45\textwidth}
        \centering
        \includegraphics[width=\textwidth]{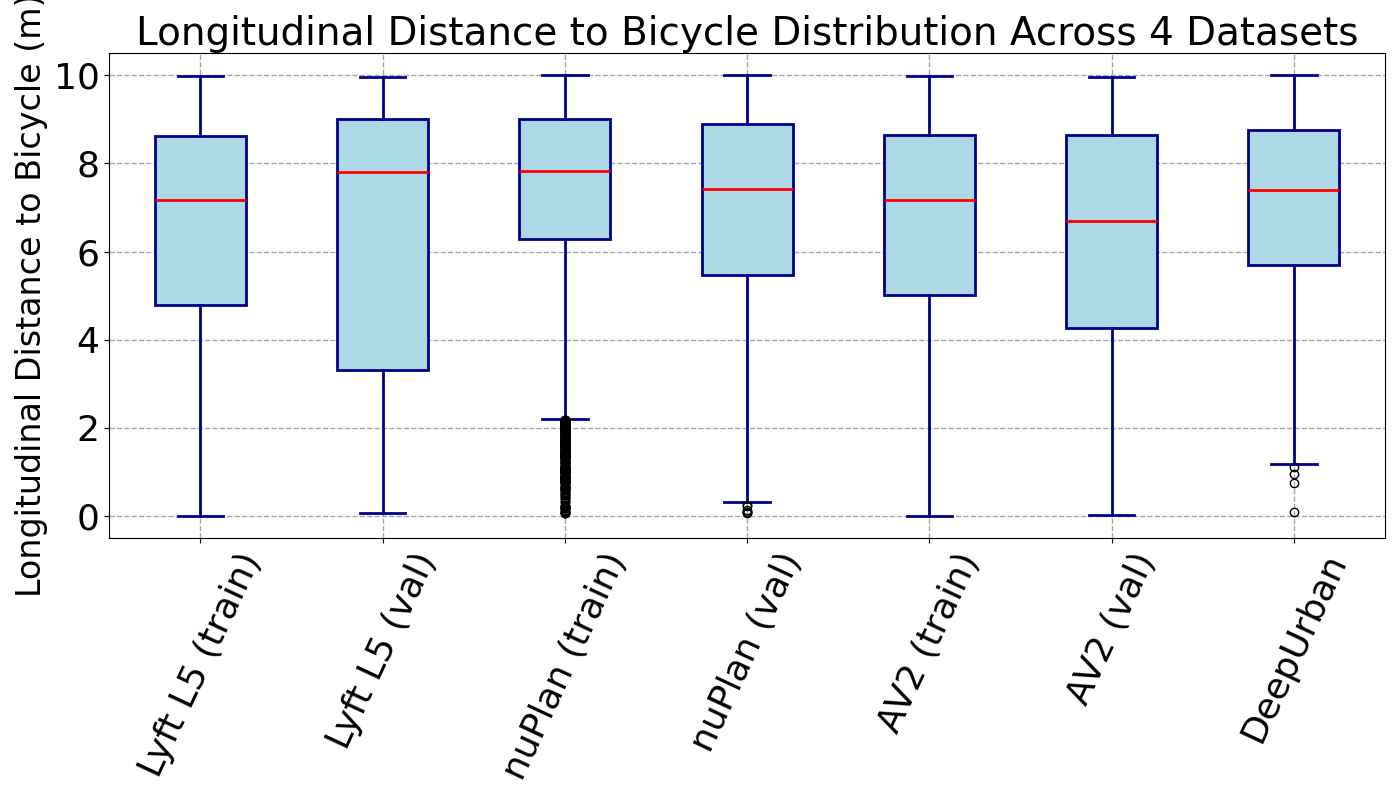}
    \end{subfigure}

    \begin{subfigure}[b]{0.45\textwidth}
        \centering
        \includegraphics[width=\textwidth]{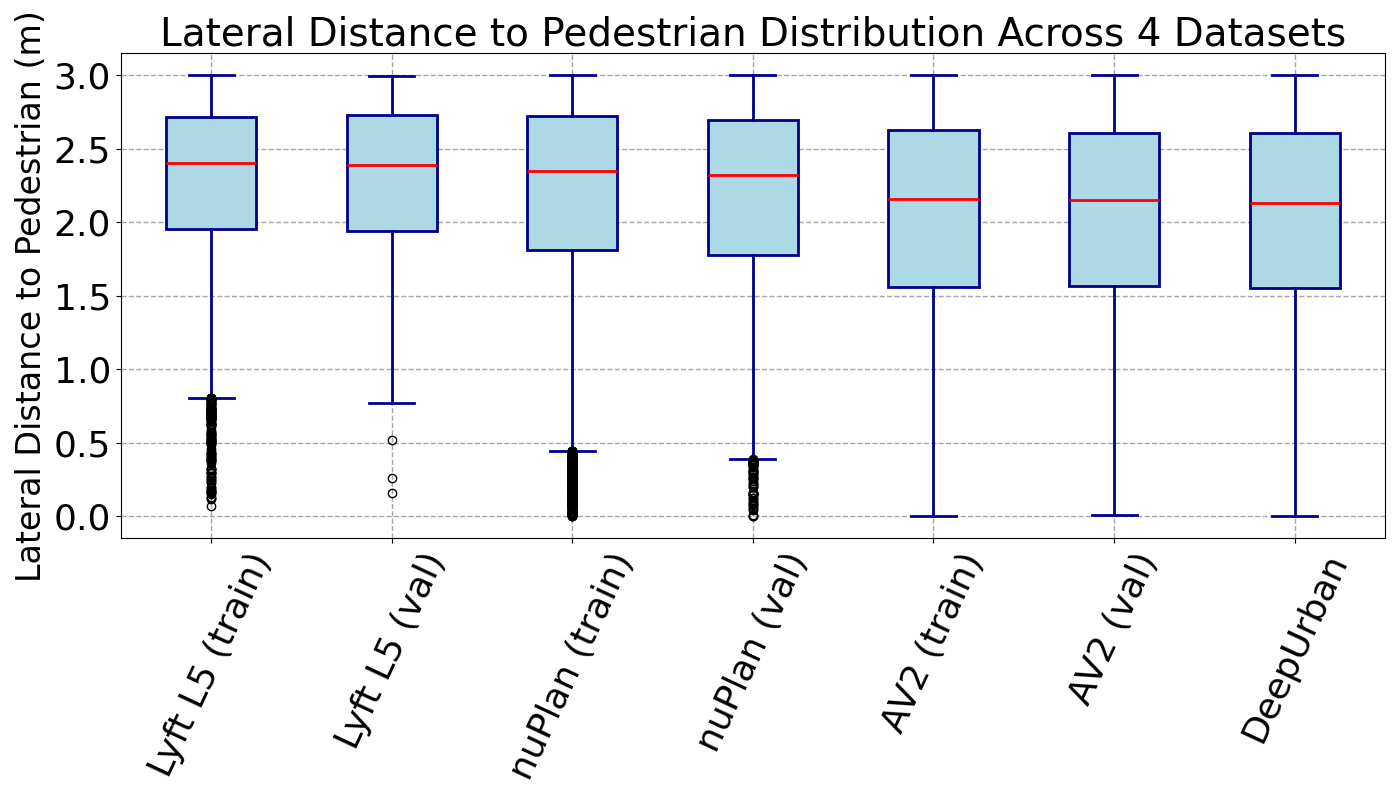}
    \end{subfigure}
    \hfill
    \begin{subfigure}[b]{0.45\textwidth}
        \centering
        \includegraphics[width=\textwidth]{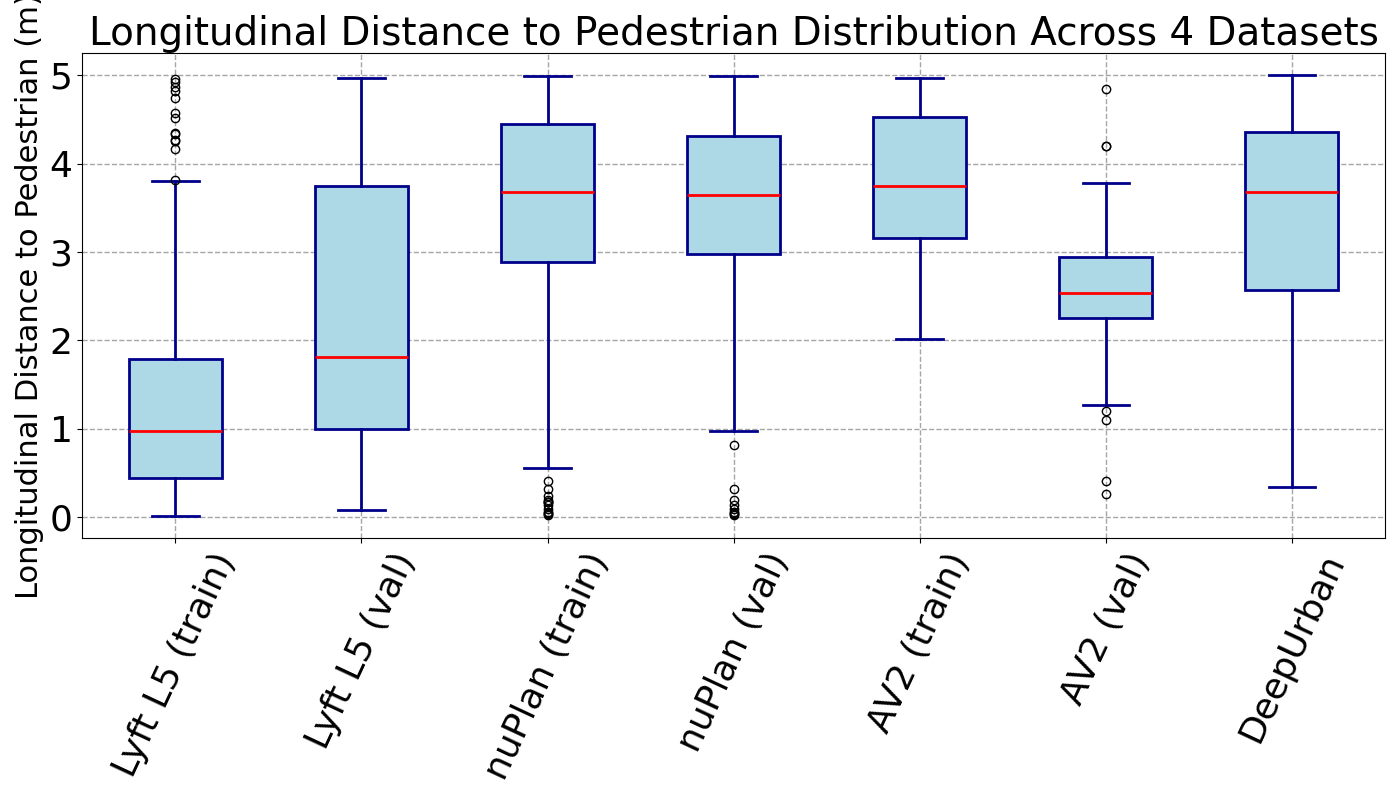}
    \end{subfigure}

    \begin{subfigure}[b]{0.45\textwidth}
        \centering
        \includegraphics[width=\textwidth]{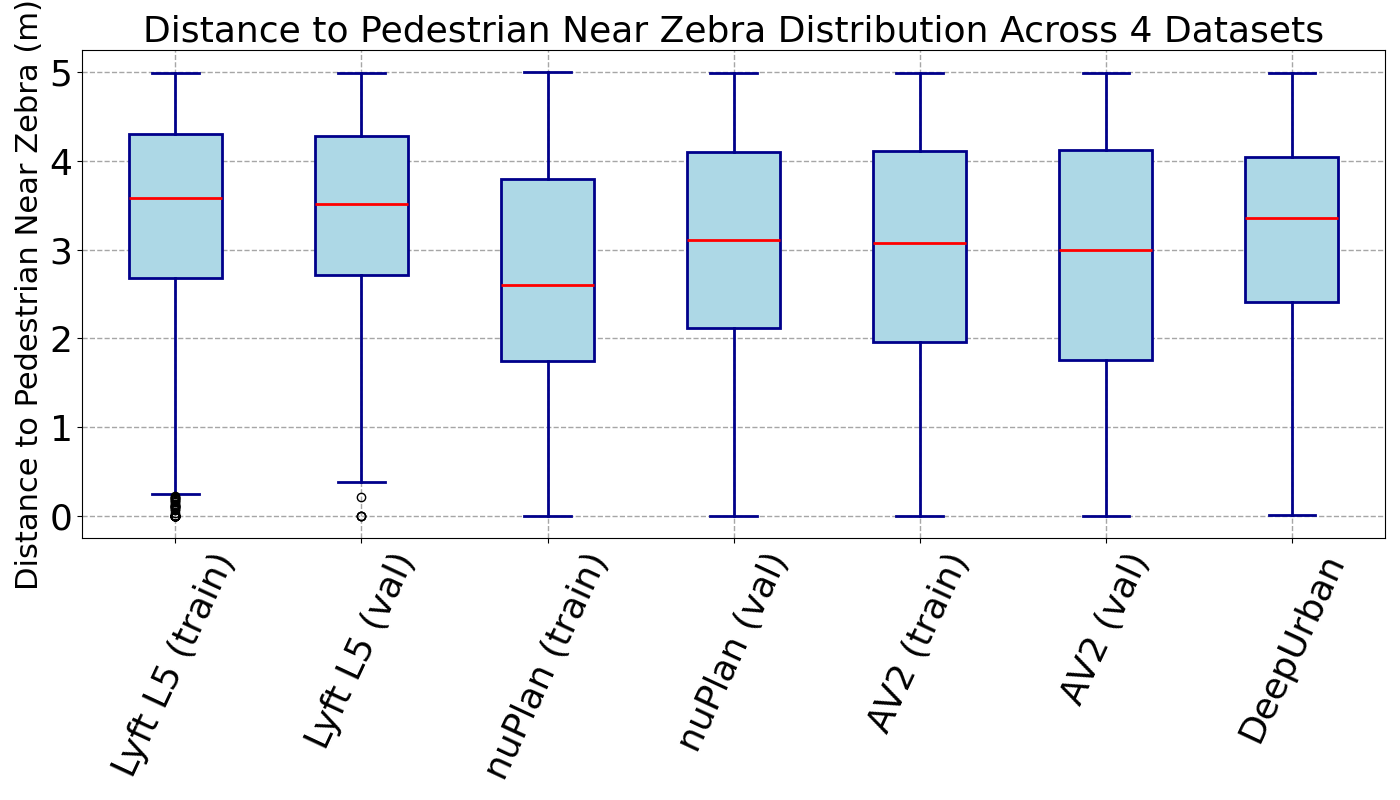}
    \end{subfigure}
    \hfill
    \begin{subfigure}[b]{0.45\textwidth}
        \centering
        \includegraphics[width=\textwidth]{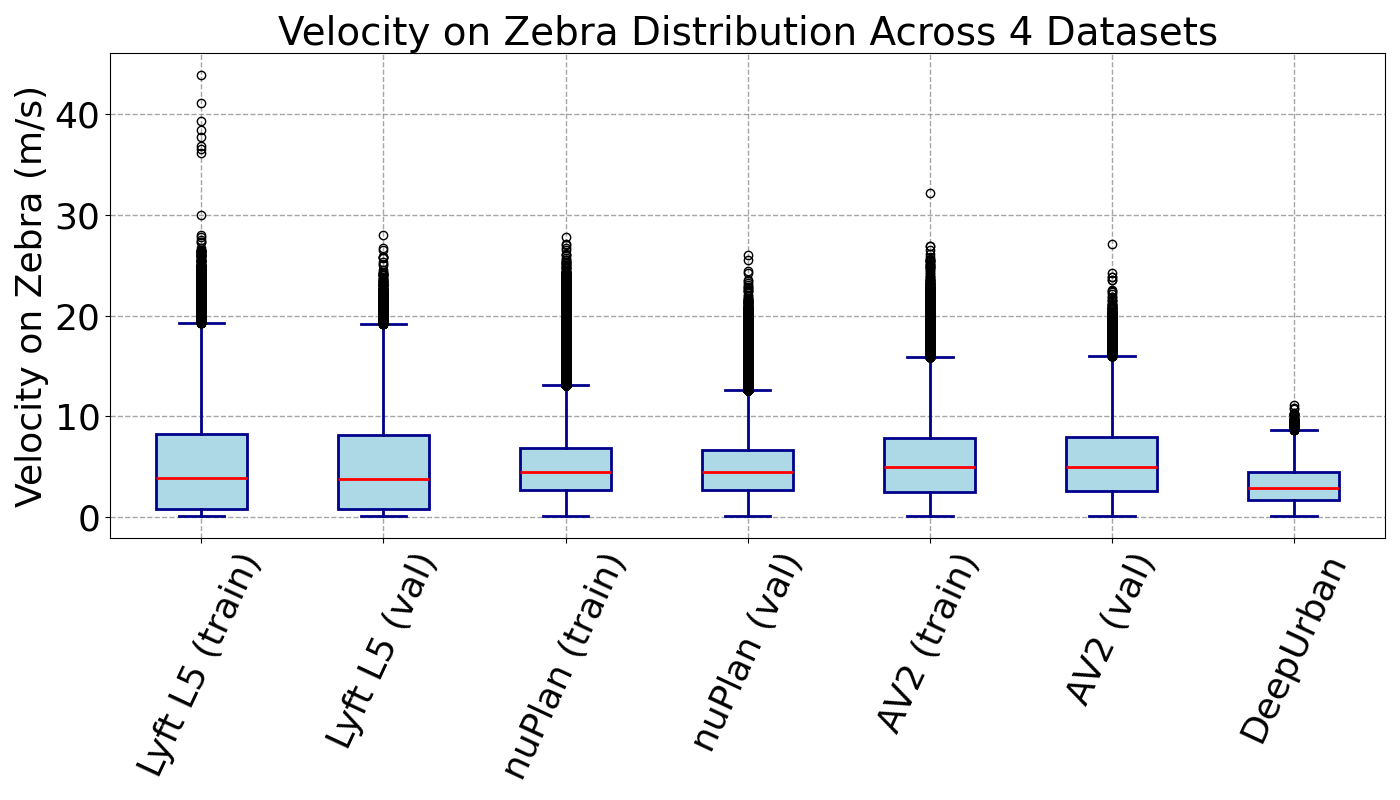}
    \end{subfigure}

    \begin{subfigure}[b]{0.45\textwidth}
        \centering
        \includegraphics[width=\textwidth]{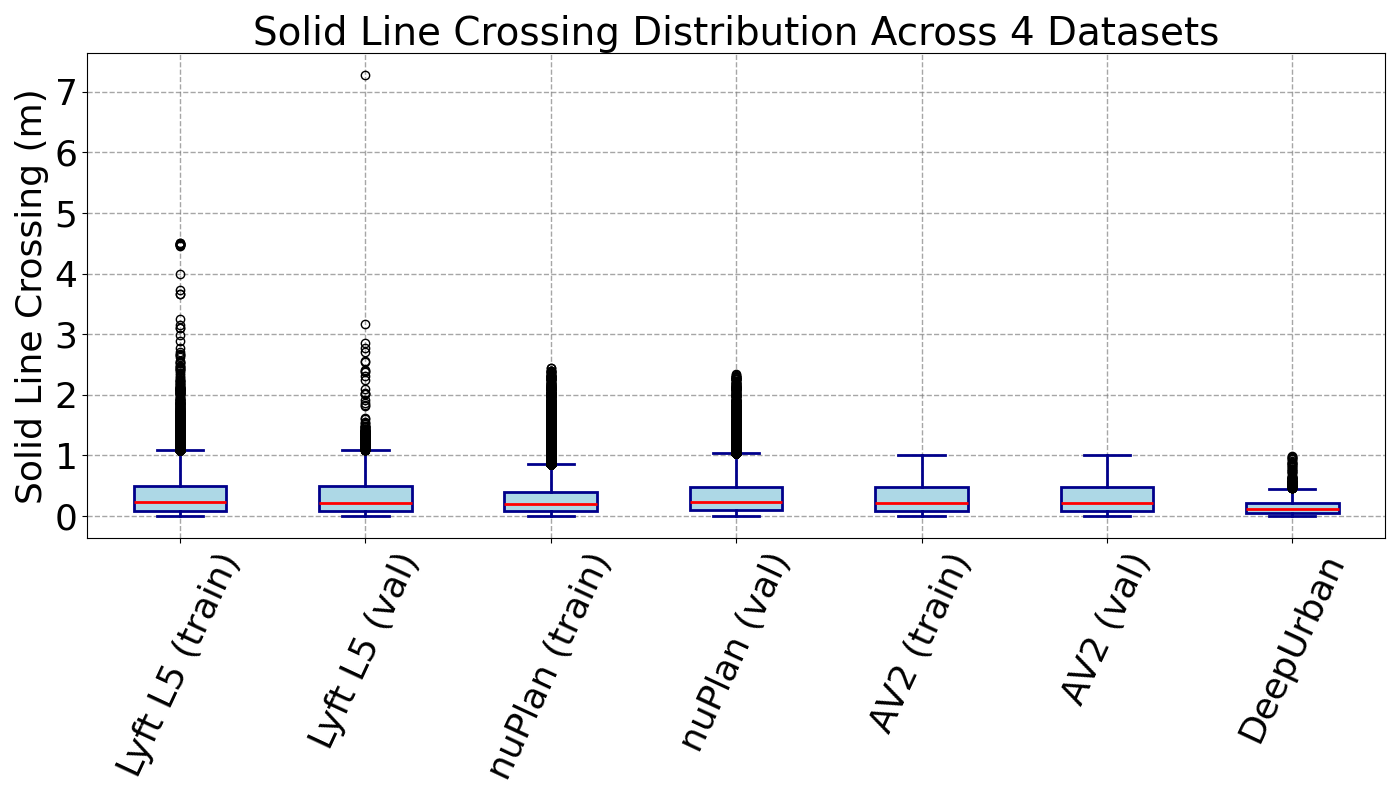}
    \end{subfigure}

    \caption{Comparison of distributions of different criteria in datasets.}
    \label{fig:comparison_datasets}
\end{figure}


\section{Results}
\label{sec:result}

The analysis focus on evaluating the performance of Lyft, nuPlan, Argoverse 2 (AV2), and DeepUrban based on the defined metrics and examining the critical agents. The distribution of each criterion is illustrated in Fig.~\ref{fig:comparison_datasets} and their corresponding medians are presented in Table~\ref{tab:medians} for more detail information.  

The velocity distribution across datasets revealed that most vehicles navigate at slower speeds, with a median velocity below 8~m/s in all datasets. The Lyft dataset showed the widest range of velocities and the highest number of outliers, reaching almost 100~m/s, likely due to noise in the data. The highest median velocity was observed in the nuPlan dataset (7.39~m/s). Values for outliers in datasets suggested that these velocities are too high to be valid, except for DeepUrban and nuPlan, where top velocities of 28~m/s and 35~m/s, respectively, are more plausible for their scenarios. 

Acceleration distributions center around 0 m/s², except for DeepUrban, which has a narrower range. Lyft has the most extreme values, suggesting that velocity outliers influence acceleration noise in both Lyft and AV2, while DeepUrban has the most precise measurements.

\begin{table}[t]
    \vspace{-0.5cm}
    \centering
    \footnotesize
    \begin{tabular}{l|c|cccccc}
    \toprule
    \textbf{Dataset} & \textbf{Total Agents} & \textbf{VEL} & \textbf{ACC} & \textbf{TTC} & \textbf{DTPNZ} & \textbf{LADTB} & \textbf{SLC} \\
    \midrule
    Lyft (train)  & 40,938,112 &  4.48\% & 9.57\% & 0.05\% & 0.00\% & 0.00\% & 0.16\% \\ 
    Lyft (val) & 4,802,359 & 4.37\%  & 9.49\% & 0.06\% & 0.00\%  & 0.00\%  & 0.15\%  \\ 
    nuPlan (train) &  12,870,111 & 1.32\% & 1.32\% & 0.00\% & 0.12\% & 0.01\% & 0.88\% \\ 
    nuPlan (val) & 2,376,986 & 0.84\% & 1.02\%  & 0.00\%  & 0.07\%  & 0.01\%  & 1.76\%  \\ 
    AV2 (train) & 8,141,840 & 6.32\%  & 11.66\%  & 0.42\%  & 0.08\%  & 0.01\%  & 0.78\%  \\ 
    AV2 (val) & 1,020,858 & 6.34\%  & 11.69\%  & 0.38\%  & 0.10\%  & 0.01\%  & 0.79\%  \\ 
    DeepUrban & 909,507 &  3.01\%&  0.00\%&  0.00\%&  0.01\% &  0.04\%& 0.04\%  \\
    \bottomrule
    \end{tabular}
    \vspace{1em}
    \caption{Comparison of critical agent amount based on different metric thresholds in proposed datasets.} 
    \label{tab:criticalagent}
        \vspace{-0.5cm}
\end{table}

The gap distribution varied significantly between datasets, with AV2 showing the highest median gap (30.21 meters), followed by Lyft (25.9 meters).
A notable observation is the difference between AV2's training and validation datasets, where agents in the validation set maintained shorter distances from each other compared to the training set. High gap values are generally safer, but shorter gaps with stationary vehicles are also safe, illustrating that gap alone is not necessarily a measure of criticality, therefore, we suggest using this metric in combination with others.
TTC is a crucial metric for assessing potential collisions. The Lyft and AV2 datasets both have a median TTC of 6~seconds, with similar distributions between their training and validation sets. nuPlan and DeepUrban have higher median TTC values of 7~seconds, indicating safer behavior among agents. Both datasets also show a similar distribution, particularly in nuPlan's validation set. Low-range outliers, particularly in nuPlan and DeepUrban, have TTC values as low as 1~second or 0.5~seconds.
The Lyft dataset demonstrates a more defensive driving style toward bicycles in LADTB, compare to all other datasets which lower quartiles fell below critical value of 1.5 meters.

\begin{figure}[t]
    \centering
    \begin{subfigure}[b]{0.48\textwidth}
        \centering
        \includegraphics[width=\textwidth]{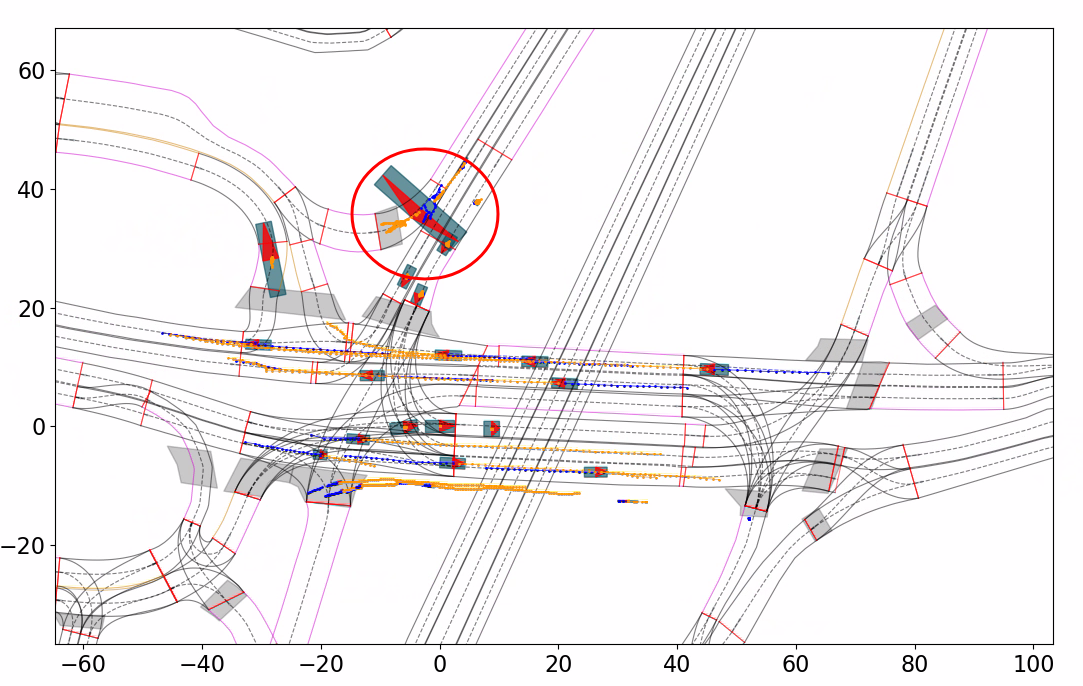}
    \end{subfigure}
    \hfill
    \begin{subfigure}[b]{0.513\textwidth}
        \centering
        \includegraphics[width=\textwidth]{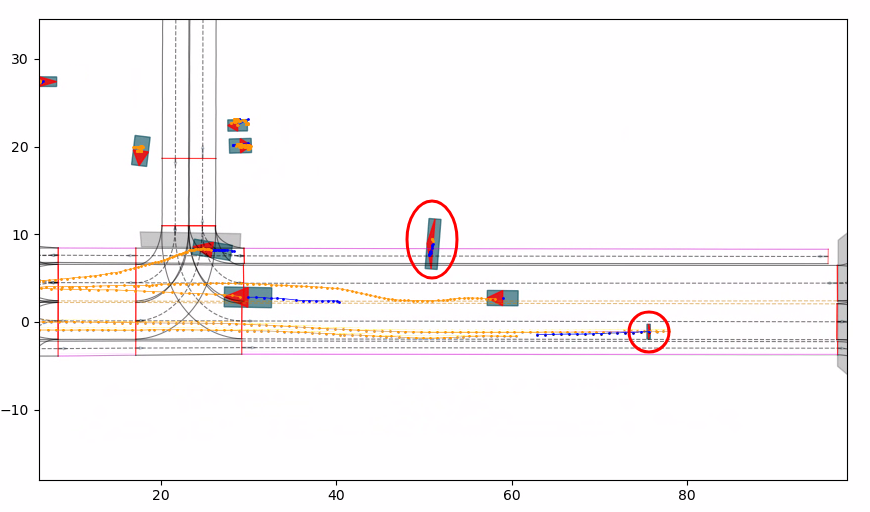}
    \end{subfigure}
    \hfill

    \vspace{-0.4cm}
    \caption{Examples of noisy data in Lyft dataset which have wrong heading angle.}
    \label{fig:critical_cases_lyft}
        \vspace{-0.4cm}
\end{figure}

Among the four metrics capturing driver-pedestrian interactions, the largest differences appear in longitudinal distance. The Lyft dataset shows more aggressive behavior, while AV2's training set exhibits more desirable behavior with most cases above 2 meters, though its validation set presents more challenging situations. DeepUrban demonstrates a more defensive driving style in VOZ, with lower median and tail values. In SLC, most agents fall between 0 and 1, showing small deviations from the center line. However, Lyft contains unrealistic cases exceeding 3 meters in SLC, likely due to errors in bounding box orientation or size estimation, as shown in
Fig.~\ref{fig:critical_cases_lyft}.

\vspace{-10pt}
\section{Conclusion}
\label{sec:conclusion}

The analysis of human driving behavior across four key datasets revealed significant differences in driving behavior and noise levels. The Lyft dataset exhibited the most noise, with extreme outliers in both velocity and acceleration, highlighting the need for thorough preprocessing to ensure accurate model training. In contrast, the DeepUrban dataset demonstrated more controlled and realistic driving behaviors, with fewer outliers and a consistently defensive driving style, making it a reliable source for analysis.
The AV2 shows a significantly higher percentage of critical TTC values, with up to 0.42\% in the training set and 0.38\% in the validation set, compared to other datasets such as Lyft and nuPlan, which mostly hover around 0\%. This higher percentage in AV2 highlights a distinct driving pattern within the dataset.
The nuPlan and AV2 datasets showed balanced driving behaviors, representing a middle ground between the more defensive DeepUrban and the outlier-heavy Lyft. The comparative analysis between Lyft and AV2 also revealed that advancements in sensors and post-processing techniques have led to noticeable reductions in noise levels over time.
Additionally, datasets captured via drones generally exhibited lower noise levels than those recorded using onboard sensors. This difference is likely due to the broader, less obstructed perspective provided by drones, whereas onboard sensors are limited by the ego vehicle's viewpoint. By employing the proposed framework, we can effectively identify abnormal behaviors across these datasets, facilitating the training of models that focus on more desirable human driving patterns.

\clearpage


\bibliography{example}  

\end{document}